\documentclass[10pt,twocolumn,letterpaper]{article}

\usepackage{cvpr}
\usepackage{times}
\usepackage{epsfig}
\usepackage{graphicx, subcaption}
\usepackage{amsmath,mathtools}
\usepackage{amssymb}
\usepackage{multirow}
\usepackage{url}
\usepackage{amssymb}
\usepackage{pifont}

\usepackage{multirow}
\usepackage{array}
\usepackage{amsbsy}
\usepackage{graphicx}
\usepackage{subcaption,adjustbox}
\usepackage{soul}
\usepackage{color}
\usepackage{breqn}
\usepackage{authblk}
\usepackage{bm}
\usepackage{amsthm}
\usepackage{siunitx}
\usepackage[linesnumbered,ruled]{algorithm2e}

\usepackage[breaklinks=true,bookmarks=false,colorlinks]{hyperref}

\cvprfinalcopy 

\def\cvprPaperID{8553} 

\begin{document}

\title{Signed Input Regularization}


\author[]{Saeid Asgari Taghanaki }
\author[]{Kumar Abhishek}
\author[]{ Ghassan Hamarneh}
\affil[]{School of Computing Science, Simon Fraser University, Canada\protect \\\texttt{\{sasgarit,kabhishe,hamarneh\}@sfu.ca}}





\maketitle
\begin{abstract}
Over-parameterized deep models usually over-fit to a given training distribution, which makes them sensitive to small changes and out-of-distribution samples at inference time, leading to low generalization performance. To this end, several model-based and randomized data-dependent regularization methods are applied, such as data augmentation, which prevents a model from memorizing the training distribution. Instead of the random transformation of the input images, we propose SIGN, a new regularization method, which modifies the input variables using a linear transformation by estimating each variable's contribution to the final prediction. Our proposed technique maps the input data to a new manifold where the less important variables are de-emphasized. To test the effectiveness of the proposed idea and compare it with other competing methods, we design several test scenarios, such as classification performance, uncertainty, out-of-distribution, and robustness analyses. We compare the methods using three different datasets and four models. We find that SIGN encourages more compact class representations, which results in the model's robustness to random corruptions and out-of-distribution samples while also simultaneously achieving superior performance on normal data compared to other competing methods. Our experiments also demonstrate the successful transferability of the SIGN samples from one model to another. 
\end{abstract}

\section{Introduction}

The classical computer vision methods generally achieve lower performance compared to the current deep neural networks~\cite{o2019deep}. This is mainly because classical methods are designed to capture human-level information from images at the expense of curtailing their level of \textit{freedom} from the beginning of the \textit{exploration} of a solution to the problem. In contrast, deep models might not be able to provide human-communicable latent information from signals, but they achieve superior performance in various machine learning tasks. However, extensive regularization of the deep networks reduces their {freedom of exploration}. 





\begin{figure}[h!]
\centering
\includegraphics[width=\linewidth]{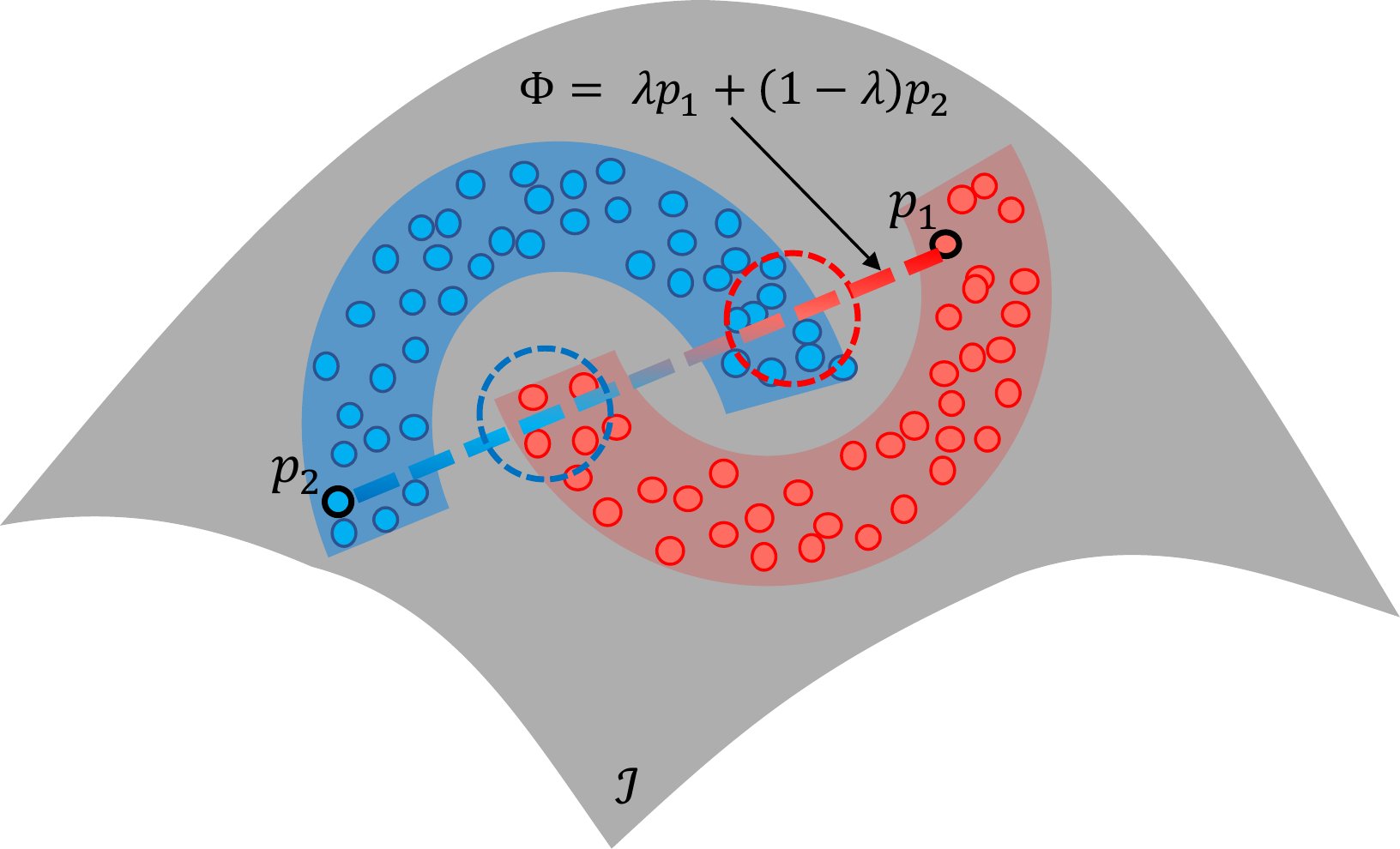}
\caption {The \textit{mixup} method leads to wrong labels for the samples on the boundaries due to random sample selection. Blue and red dots represent samples from two different classes. The dashed line represents the interpolation function ($\lambda(p_{1}) + (1-\lambda)p_{2}$) for both images and labels, which was proposed by the authors~\cite{zhang2017mixup} for the mixing step. The samples inside the red dotted circle should be assigned a blue label; however, due to random sampling, they get a red label, and vice-versa for the blue dotted circle. $\mathcal{J}$ represents the image space.}
\label{mixup}
\end{figure}

Several different regularization techniques have been successfully applied to over-parameterized models to reduce over-fitting and to generalize to unseen data, including the $L_2$ norm penalty for model parameters, weight decay~\cite{nowlan1992simplifying}, early stopping, and dropout~\cite{srivastava2014dropout}. Although initially not introduced as a regularization method, batch normalization~\cite{ioffe2015batch} performs regularization by considering fluctuations within mini-batches, albeit with a generalization performance of deep neural networks that often depends on the size of the mini-batch~\cite{keskar2016large}. Similarly, stochastic gradient descent (SGD)~\cite{sutskever2013importance} can also be interpreted as a regularized gradient descent impaired by noisy gradients. However, SGD's fluctuations do not lead to a smooth convergence. Other types of regularization are data-dependent, such as traditional data augmentation techniques (rotation, flipping, etc.)~\cite{lecun1998gradient,simonyan2014very}, AutoAugment~\cite{cubuk2018autoaugment}, and the mixup method~\cite{zhang2017mixup}. The regularization properties of mixup were studied by Guo et al.~\cite{guo2019mixup}. Although a few works have attempted to improve the mixup method~\cite{shimada2019data,mai2019metamixup}, besides the smoothed decision boundaries which can cause a model to be highly under-confident, the random sample selection step of mixup may produce wrong labels, as shown in Figure~\ref{mixup}, thus failing to generalize well. Another issue with the mixup method is that the linear blending results in implausible images. As we detail later, our proposed method modifies the input to generate new samples as well, but it does so while avoiding the drawbacks of the linear blending.

\textbf{Adversarial training} is another data-dependent regularization approach~\cite{goodfellow2014explaining,roth2019adversarial} in which an adversarially perturbed version of data is used as a type of augmentation. However, an adversarially perturbed image contains \textit{perceptible} information about the correct class and \textit{imperceptible} information about a wrong or a random label. Ideally, a model should learn to cancel out the deliberate implicit patterns specific to the wrong class, but with an arbitrary level of perturbation (which is the case in adversarial training), a model might get biased towards an implausible distribution. In other words, from an adversarial attack perspective, it is ideal to destroy a well-distributed manifold. Therefore, using adversarial training as data augmentation for regularization can skew the feature space towards an arbitrary class distribution. 

In contrast to adversarial training that perturbs an input image to effectuate a wrong prediction, Taghanaki et al.~\cite{taghanakitransfer} proposed a gradient based method that perturbs the input image by transforming it to a new space to effectuate a more accurate prediction. However, since their method relies on the true labels, which are not available at the inference time, they leverage a second model to learn the mapping between the input and the corresponding transformed images, and show an improvement in performance with a computational complexity trade-off. As we describe later, our proposed method also generates new perturbed images that lead to improved predictions, however, it does not require ground truth labels for mapping. 

\textbf{Relation to causality.} Understanding the effect of input variables on the prediction function’s output is essentially a causality problem. Recently, several approaches have successfully applied reverse gradients towards the input to decode, and in particular, to \textit{visualize} the causal effect of input variables on the overall final prediction~\cite{bach2015pixel,selvaraju2017grad,smilkov2017smoothgrad,sundararajan2017axiomatic}. However, they have not studied the causal influence of each single input variable towards a better prediction performance. In this paper, we attempt to answer the question: ``can the casual influence of each variable in the input manifold be modified to limit the number of input variables by discarding the less effective ones?". If a model can successfully capture only the truly correlated variables in the input space to the labels, then the model should be robust to manifold shifts and out-of-distribution samples~\cite{bengio2019meta} without the need to perform explicit regularization. Leveraging prior knowledge, such as location information of target objects, might help regularize the model and steer it to focus only on the relevant variables in the input, which has motivated recent works on attention priors~\cite{yan2019melanoma,zhao2019retinal}. However, such location labels are not always provided and, similar to over-regularization, excessive prior knowledge might restrict the model's {exploration} of the data and miss other more discriminative latent information.

To facilitate the model's exploration with only relevant input variables, we seek a transformation that projects the input to a new space where the important variables are emphasised. Therefore, we adopt the push-forward approach~\cite{jost2008riemannian,lee2013smooth} to approximate the changes in the output for each input variable. In contrast to methods that limit the {exploration} of the model by turning off the neurons and scaling the model parameters, such as dropout and batch normalization, and inspired by the adversarial training concept, we introduce a new method called \textbf{S}igned \textbf{I}nput Re\textbf{g}ularizatio\textbf{n} (SIGN). 

The SIGN method re-weights the input (pixels in the image space, $\mathcal{J}$) such that an input sample $p \in \mathbb{R}^+$ is transformed to $p + \delta \in \mathbb{R}$. The variables $\in \mathbb{R}^- $ are discarded by a model with ReLU non-linearity. Thus the model focuses on fewer input variables.

In this paper, we make the following {contributions}:
\begin{itemize}
    \item We introduce a new Jacobian-based regularization method which facilitates the model's exploration by reducing the variables in the input.
    \item We show how the push-forward and manifold mapping concepts can be used for the linear estimation of input using a layer's output in a model.
    \item Our proposed method achieves significantly better results compared to traditional data augmentation and the mixup regularization schemes. We also discuss and demonstrate the potential shortcomings of the mixup method.
\end{itemize}

\section{Method}

\begin{figure}[h!]
\centering
\includegraphics[width=\linewidth]{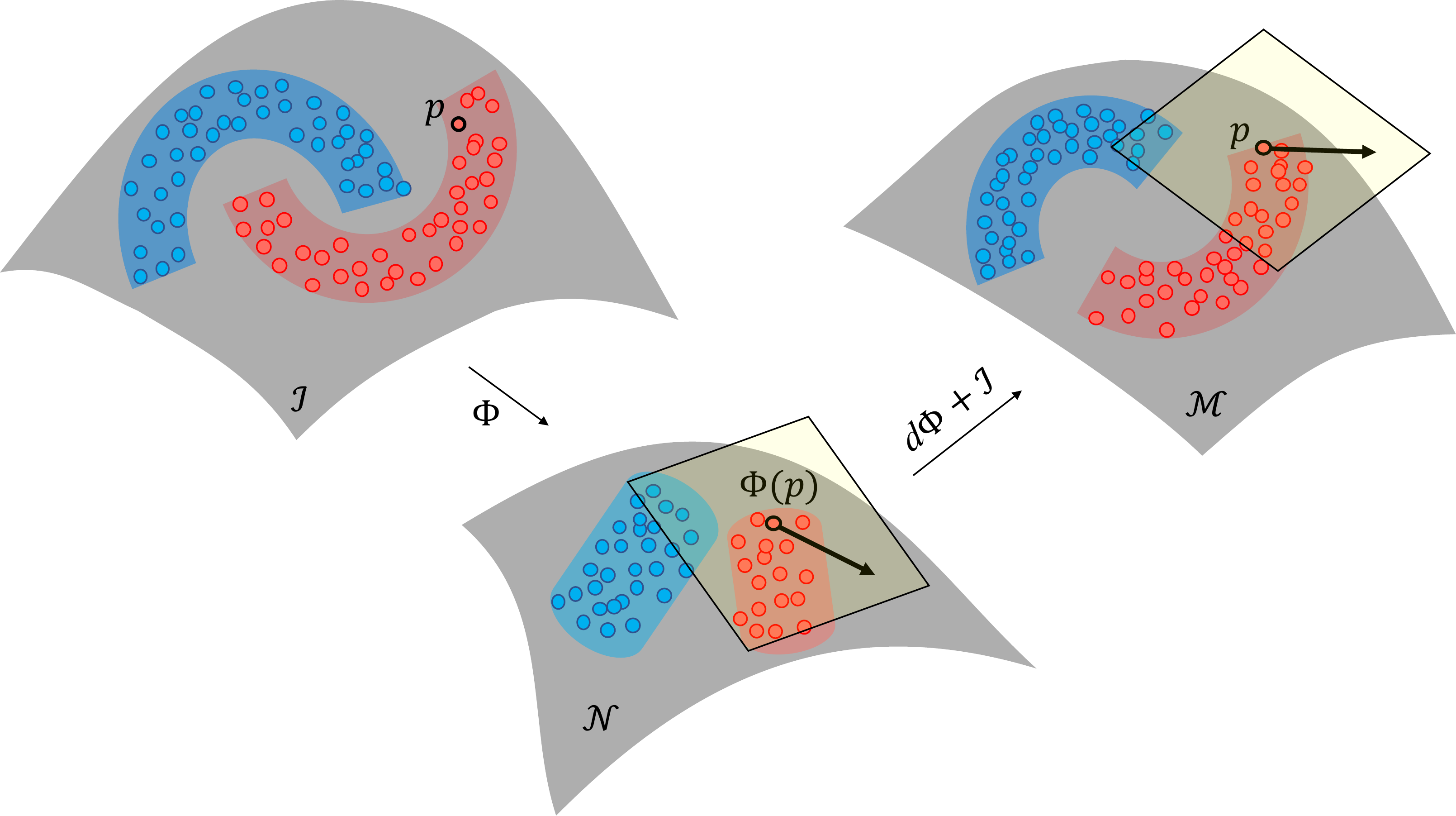}
\caption {Visualization of the push-forward map from the tangent space in $\mathcal{N}$ to the tangent space in $\mathcal{M}$. Compared to the mixup method (Figure~\ref{mixup}), which transforms the sample in the same manifold using a linear combination, our proposed method moves the samples to a new manifold. $\mathcal{J}$ represents the image space.}
\label{pushf-fig}
\end{figure}

Given a training sample $p$, the goal is to transform it to a new space. Let $\Phi: \mathcal{N}\rightarrow\mathcal{M}$ be a map between manifolds $\mathcal{N}$ and $\mathcal{M}$. Then, the derivative $d\Phi(p)$, the derivative of $\Phi$ at point $p$ in the tangent space of $\mathcal{N}$ is the best linear approximation of $\Phi$ near $p$ in the tangent space of $\mathcal{M}$. Therefore, $d\Phi(p)$ pushes tangent vectors on $\mathcal{N}$ forward to tangent vectors on $\mathcal{M}$~\cite{lee2013smooth}, as shown in Figure~\ref{pushf-fig}.

\subsection{The Push Forward Map}\label{pushf}
Let $\Phi: \mathbb{R}^{n} \rightarrow \mathbb{R}^{m}$ be a smooth function ($C^{\infty}$) and let $p \in \mathbb{R}^{n}$. The goal is to define the push forward map~\cite{jost2008riemannian,lee2013smooth} as 


\begin{equation}
    \Phi_{*, p}: T_{p} \mathbb{R}^{n} \rightarrow T_{\Phi(p)} \mathbb{R}^{m}.
\end{equation}

\smallskip

Let $X_{p} \in T_{p} \mathbb{R}^{n}$ be a tangent vector (in tangent space $T$) which by definition is completely determined by its action on $C^{\infty}(p)$. In order that $\Phi_{*, p}\left(X_{p}\right)$ to be a tangent vector at $q=\check{\Phi}(p) \in \mathbb{R}^{m}$, we need to define 

\begin{equation}
    \Phi_{*, p}\left(X_{p}\right)(g) \quad \forall g \in C^{\infty}(q).
\end{equation}

\smallskip

Let $g \in C^{\infty}(q)$ be a function in the space of $\Phi$. The composition $g \circ \Phi: \mathbb{R}^{n} \rightarrow \mathbb{R}$ is a smooth function on an open subset of $\mathbb{R}^{n}$ which contains $p$, and thus $X_{p}(g \circ \Phi)$ is well defined. Let $\Phi: \mathbb{R}^{n} \rightarrow \mathbb{R}^{m}$ be a smooth function and let $q = \Phi(p)$. Given $X_{p} \in T_{p} \mathbb{R}^{n} \text { , } \Phi_{*, p}\left(X_{p}\right): C^{\infty}(q) \rightarrow \mathbb{R}$ is defined by 

\begin{equation}\label{eq3}
    \Phi_{*, p}\left(X_{p}\right)(g)=X_{p}(g \circ \Phi) \quad \forall g \in C^{\infty}(q)
\end{equation} 

\smallskip

Equation~\ref{eq3} can be written using the coordinate bases. Let $\beta=\left\{\left.\partial_{x^{i}}\right|_{p}\right\}_{1 \leq i \leq n}$ be the coordinate basis for $T_{p} \mathbb{R}^{n}$ and let $\gamma= \left\{\left.\partial_{y^{a}}\right|_{q}\right\}_{1 \leq a \leq m}$
be the coordinate basis for $T_{q} \mathbb{R}^{m}$. If $X_{p} \in T_{p} \mathbb{R}^{n}$ is given by 

\smallskip

\begin{equation}
    X_{p}=\left.\sum_{i=1}^{n} \xi^{i} \partial_{x^{i}}\right|_{p}, \quad \xi^{i} \in \mathbb{R},
\end{equation}

then,
\begin{equation} \label{eq5}
    \Phi_{*, p} X_{p}=\left.\sum_{a=1}^{m}\left(\left.\sum_{i=1}^{n} \xi^{i} \frac{\partial \Phi^{a}}{\partial x^{i}}\right|_{p}\right) \frac{\partial}{\partial y^{a}}\right|_{q}.
\end{equation}

\smallskip

Therefore, the matrix representation of the linear transformation $\Phi_{*, p}: T_{p} \mathbb{R}^{n} \rightarrow T_{q} \mathbb{R}^{m}$ in the basis $\beta=\left\{\left.\partial_{x^{i}}\right|_{p}\right\}_{1 \leq i \leq n}$ for $T_{p} \mathbb{R}^{n}$ and the basis $\gamma=\left\{\left.\partial_{y^{a}}\right|_{q}\right\}_{1 \leq a \leq m}$ for $T_{\Phi(p)} \mathbb{R}^{m}$ is given by the Jacobian

\smallskip

\begin{equation}
    \left[\Phi_{*, p}\right]=\left.\frac{\partial \Phi^{a}}{\partial x^{i}}\right|_{p}
\end{equation}

\smallskip

\noindent\textbf{\textit{Proof.}} If we write the matrix representation of $\Phi_{*, p}$ using the matrix $\left(J_{i}^{a}\right) \in M_{m \times n}(\mathbb{R})$, we get

\begin{equation}
    \Phi_{*, p}\left(\left.\partial_{x^{i}}\right|_{p}\right)=\left.\sum_{a=1}^{m} J_{i}^{a} \partial_{y^{a}}\right|_{q}.
\end{equation}

\smallskip

The matrix $J_{i}^{a}$ can then be calculated using Equation~\ref{eq5}, and therefore

\begin{equation}
    \Phi_{*, p}\left(\left.\partial_{x^{i}}\right|_{p}\right)=\left.\left.\sum_{a=1}^{m} \frac{\partial \Phi^{a}}{\partial x^{i}}\right|_{p} \partial_{y^{a}}\right|_{q}
\end{equation}

\begin{equation*}
    \implies \left[\Phi_{*, p}\right]=\left.\frac{\partial \Phi^{a}}{\partial x^{i}}\right|_{p}.   \qedhere
\end{equation*}

Therefore, the map $\Phi_{*, p}$ can be thought of as a matrix of partial derivatives:

\smallskip

\begin{equation}
    \Phi_{*,p}=\left(\begin{array}{cccc}{\frac{\partial \phi^{1}}{\partial x^{1}}} & {\frac{\partial \phi^{1}}{\partial x^{2}}} & {\cdots} & {\frac{\partial \phi^{1}}{\partial x^{m}}} \\ {\frac{\partial \phi^{2}}{\partial x^{1}}} & {\frac{\partial \phi^{2}}{\partial x^{2}}} & {\cdots} & {\frac{\partial \phi^{2}}{\partial x^{m}}} \\ {\vdots} & {\vdots} & {\ddots} & {\cdots} \\ {\frac{\partial \phi^{n}}{\partial x^{1}}} & {\frac{\partial \phi^{n}}{\partial x^{2}}} & {\cdots} & {\frac{\partial \phi^{n}}{\partial x^{m}}}\end{array}\right).
\end{equation}

\smallskip

\subsection{Iterative SIGN Method}
Given a trained model, the goal is to transform the input with a linear estimation using the model's $l^{th}$ layer under the push-forward definition, which is realized by the Jacobian, as explained in Section~\ref{pushf}. Not only the Jacobian provides the linear estimation of the input, it also highlights the variables in the input which contribute the most to the output of the layer. The input variables with negative sign are discarded after they are transferred to manifold $\mathcal{M} \in \mathbb{R}$ because the following ReLU activation function does not let them pass on to the subsequent layers. 

In order to emphasize the input variable discarding step, we use a similar approach to the momentum method~\cite{polyak1964some}, which is a technique for accelerating gradient descent algorithms by accumulating a velocity vector in the gradient direction of the loss function across iterations. Our iterative process is summarized in Algorithm~\ref{signn}. 

\begin{algorithm}[] 
 \SetKwInOut{Input}{Input}
 \SetKwInOut{Output}{Output}
 \Input{A well-trained model $\Phi$, a real sample $p$, and number of iterations $K$}
 \Output{Transformed input sample $p^*$}
 $p_{0}^* = p$\;
 \For{$k = 0$ \textrm{to} $K-1$} {
    $\delta_{k+1} = \sum_{a} \left.\frac{\partial \Phi^{a}}{\partial x^{i}}\right|_{p}$ \;
    $p_{k+1}^* = p^* + \delta_{k+1}$
}
$p^* = p_{K}^*$\;
\Return{$p^*$}
\caption{Iterative SIGN}
\label{signn}
\end{algorithm}

\begin{table*}[!ht]
\centering
\caption{Class-wise and overall classification performances of different methods on CIFAR-10 using MobileNetV2.}
\resizebox{\textwidth}{!}{%
\begin{tabular}{clccccccccccc}
\hline

Model  & Method  & airplane  & automobile  & bird  & cat  & deer  & dog  & frog  & horse  & ship  & truck  & Mean  \\ \hline
\multirow{3}{*}{\begin{tabular}[c]{@{}c@{}}MobileNetV2\\\cite{sandler2018mobilenetv2}\end{tabular}}
& Classical~\cite{simonyan2014very}  & $0.834$  & $0.940$  & $0.850$  & $0.691$  & $0.858$  & $0.758$  & $0.914$  & $0.875$  & $0.909$  & $0.921$  & $0.8550$  \\
& mixup~\cite{zhang2017mixup}  & $\textbf{0.870}$  & $0.902$  & $\textbf{0.861}$  & $\textbf{0.793}$  & $0.810$  & $0.716$  & $0.886$  & $\textbf{0.884}$  & $0.935$  & $0.927$  & $0.8584$  \\
& SIGN (proposed)  & $0.868$  & $\textbf{0.931}$  & $0.806$  & $0.717$  & $\textbf{0.864}$  & $\textbf{0.763}$  & $\textbf{0.936}$  & $0.875$  & $\textbf{0.941}$  & $\textbf{0.936}$  & $\textbf{0.8637}$  \\ \hline

\hline
{BasicCNN}
& DeepAugment~\cite{deepaugment}  & $\textbf{0.885}$  & $0.930$  & $0.794$  & $\textbf{0.733}$  & $\textbf{0.911}$  & $0.746$  & $0.893$  & $\textbf{0.915}$  & $\textbf{0.924}$  & $0.917$  & $0.8648$  \\
(Section~\ref{sec:experiments})  & DeepAugment + SIGN  & $0.866$  & $\textbf{0.939}$  & $\textbf{0.819}$  & $0.728$  & $0.867$  & $\textbf{0.816}$  & $\textbf{0.920}$  & $0.898$  & $0.919$  & $\textbf{0.919}$  & $\textbf{0.8691}$  \\ \hline
\end{tabular}
}
\label{tab:classification}
\end{table*}

{ \renewcommand{\arraystretch}{1.3}

\begin{table*}[!htbp]
\centering

\caption{Aleatoric uncertainty analysis of different methods on CIFAR-10 dataset.}
\resizebox{\textwidth}{!}{%
\begin{tabular}{ccccccccccccc}
\hline

Method  & Metric  & airplane  & automobile  & bird  & cat  & deer  & dog  & frog  & horse  & ship  & truck  & Mean  \\ \hline
\multirow{3}{*}{Classical~\cite{simonyan2014very}} & Overall accuracy & $0.851$  & $0.930$  & $0.716$  & $0.721$  & $0.842$  & $0.631$  & $\textbf{0.934}$  & $0.864$  & $0.816$  & $\textbf{0.929}$  & $0.8234$  \\ \cline{2-13}

& \begin{tabular}[c]{@{}c@{}}$p \leq 0.5$\\(\# images)  \end{tabular}  & \begin{tabular}[c]{@{}c@{}}$0.4325$\\$(17)$  \end{tabular}  & \begin{tabular}[c]{@{}c@{}}$0.4081$\\$(6)$  \end{tabular}  & \begin{tabular}[c]{@{}c@{}}$0.4370$\\$(31)$  \end{tabular}  & \begin{tabular}[c]{@{}c@{}}$\textbf{0.4414}$\\$(40)$  \end{tabular}  & \begin{tabular}[c]{@{}c@{}}$0.4346$\\$(30)$  \end{tabular}  & \begin{tabular}[c]{@{}c@{}}$0.4449$\\$(41)$  \end{tabular}  & \begin{tabular}[c]{@{}c@{}}$0.4402$\\$(9)$  \end{tabular}  & \begin{tabular}[c]{@{}c@{}}$0.4435$\\$(19)$  \end{tabular}  & \begin{tabular}[c]{@{}c@{}}$0.4244$\\$(12)$  \end{tabular}  & \begin{tabular}[c]{@{}c@{}}$0.4102$\\$(7)$  \end{tabular}  & \begin{tabular}[c]{@{}c@{}}$0.4370$\\$(212)$  \end{tabular}  \\

& [uncertainty]  & $[6.5198]$  & $[5.6784]$  & $[7.1620]$  & $[7.3614]$  & $[7.3126]$  & $[7.5829]$  & $[7.0295]$  & $[7.5014]$  & $[5.8672]$  & $[6.1368]$  & $[7.1265]$  \\ \hline

\multirow{3}{*}{mixup~\cite{zhang2017mixup}} & Overall accuracy & $\textbf{0.866}$  & $0.900$  & $0.707$  & $0.625$  & $0.819$  & $0.689$  & $0.913$  & $0.912$  & $0.910$  & $0.919$  & $0.8260$  \\ \cline{2-13}

& \begin{tabular}[c]{@{}c@{}}$p \leq 0.5$\\(\# images)  \end{tabular}  & \begin{tabular}[c]{@{}c@{}}$0.3326$\\$(30)$  \end{tabular}  & \begin{tabular}[c]{@{}c@{}}$0.4394$\\$(15)$  \end{tabular}  & \begin{tabular}[c]{@{}c@{}}$0.4136$\\$(46)$  \end{tabular}  & \begin{tabular}[c]{@{}c@{}}$0.4110$\\$(93)$  \end{tabular}  & \begin{tabular}[c]{@{}c@{}}$\textbf{0.4348}$\\$(50)$  \end{tabular}  & \begin{tabular}[c]{@{}c@{}}$0.4175$\\$(59)$  \end{tabular}  & \begin{tabular}[c]{@{}c@{}}$0.4296$\\$(25)$  \end{tabular}  & \begin{tabular}[c]{@{}c@{}}$0.4053$\\$(27)$  \end{tabular}  & \begin{tabular}[c]{@{}c@{}}$0.4314$\\$(12)$  \end{tabular}  & \begin{tabular}[c]{@{}c@{}}$\textbf{0.4376}$\\$(6)$  \end{tabular}  & \begin{tabular}[c]{@{}c@{}}$0.4123$\\$(363)$  \end{tabular}  \\

& [uncertainty]  & $[3.4691]$  & $[4.5749]$  & $[4.2924]$  & $[4.6358]$  & $[4.3798]$  & $[4.6574]$  & $[4.3065]$  & $[4.5253]$  & $[4.6718]$  & $[4.4918]$  & $[4.4295]$  \\ \hline

\multirow{3}{*}{SIGN (proposed)} & Overall accuracy  & $0.864$  & $\textbf{0.949}$  & $\textbf{0.807}$  & $\textbf{0.725}$  & $\textbf{0.849}$  & $\textbf{0.764}$  & $0.895$  & $\textbf{0.918}$  & $\textbf{0.926}$  & $0.894$  & $\textbf{0.8591}$  \\ \cline{2-13}

& \begin{tabular}[c]{@{}c@{}}$p \leq 0.5$\\(\# images)  \end{tabular}  & \begin{tabular}[c]{@{}c@{}}$\textbf{0.4292}$\\$(\textbf{2})$  \end{tabular}  & \begin{tabular}[c]{@{}c@{}}$\textbf{0.4571}$\\$(\textbf{1})$  \end{tabular}  & \begin{tabular}[c]{@{}c@{}}$\textbf{0.4468}$\\$(\textbf{4})$  \end{tabular}  & \begin{tabular}[c]{@{}c@{}}$0.4199$\\$(\textbf{6})$  \end{tabular}  & \begin{tabular}[c]{@{}c@{}}$0.4275$\\$(\textbf{2})$  \end{tabular}  & \begin{tabular}[c]{@{}c@{}}$\textbf{0.4576}$\\$(\textbf{3})$  \end{tabular}  & \begin{tabular}[c]{@{}c@{}}$\textbf{0.4535}$\\$(\textbf{3})$  \end{tabular}  & \begin{tabular}[c]{@{}c@{}}$\textbf{0.4574}$\\$(\textbf{3})$  \end{tabular}  & \begin{tabular}[c]{@{}c@{}}$\textbf{0.4984}$\\$(\textbf{1})$  \end{tabular}  & \begin{tabular}[c]{@{}c@{}}$0.4322$\\$(\textbf{1})$  \end{tabular}  & \begin{tabular}[c]{@{}c@{}}$\textbf{0.4428}$\\$(\textbf{26})$  \end{tabular}  \\

& [uncertainty]  & $[6.4461]$  & $[8.4259]$  & $[6.1156]$  & $[5.9603]$  & $[5.7440]$  & $[6.7875]$  & $[6.4557]$  & $[6.8555]$  & $[8.9375]$  & $[8.4732]$  & $[6.5668]$  \\ \hline

\end{tabular}
}
\label{tab:classificationalea}
\end{table*}
}

\section{Experiments}   \label{sec:experiments}

We evaluate our proposed method on three datasets: CIFAR-10~\cite{krizhevsky2009learning}, Tiny ImageNet~\cite{chrabaszcz2017downsampled}, and 2D RGB skin lesion classification dataset from the 2017 IEEE International Skin Imaging Collaboration (ISIC) ISBI Challenge~\cite{codella2018skin}, and four different deep neural network architectures: MobileNetV2~\cite{sandler2018mobilenetv2}, Inception-ResNet-v2~\cite{szegedy2017inception}, NASNetMobile~\cite{zoph2018learning}, and a simple small network (called BasicCNN) with the following architecture to show the effectiveness of the proposed method regardless of the complexity of the architecture. 

\smallskip
\noindent\texttt{\small{[Conv2D_a $\rightarrow$ ReLu $\rightarrow$ Conv2D_a $\rightarrow$ ReLu $\rightarrow$ MaxPool2D $\rightarrow$ Dropout] $\rightarrow$ [Conv2D_b $\rightarrow$ ReLu $\rightarrow$ Conv2D_b $\rightarrow$ ReLu $\rightarrow$ MaxPool2D $\rightarrow$ Dropout] $\rightarrow$ [FC512 $\rightarrow$ ReLu $\rightarrow$ Dropout]}},
\smallskip

\noindent where \texttt{Conv2D_a} and \texttt{Conv2D_b} layers have 32 and 64 filter channels each. All the \texttt{Conv2D} layers use $3\times3$ kernels, all the \texttt{MaxPool2D} layers use $2\times2$ kernels, and all the dropout layers have a drop probability of $0.3$.

We compare our proposed SIGN method with classical data augmentation strategies (horizontal and vertical flipping, shifting, and rotations) on normal data, the mixup transformed samples, and DeepAugment~\cite{deepaugment}, the Bayesian version of the AutoAugment~\cite{cubuk2018autoaugment} methods. We train all the models with a batch size of 128 except for the Inception-ResNet-v2 in the skin lesion classification experiment, for which we set the batch size to 32. 
We train all the models except BasicCNN for 100 epochs and select the epochs corresponding to the highest accuracies on the validation sets.
 The BasicCNN model, which was used to test DeepAugment was trained for 200 epochs. 

We set $K$ (the number of iterations) for iterative SIGN (Algorithm~\ref{signn}) to 50 and 100, thereby augmenting the training set with two different versions. For the source manifold, we use the output of the last layer (before the logits) of the model to estimate the target manifold which is the input variables' space. 

In the following subsections, we start with a general classification performance analysis in subsection~\ref{classification-performance}. In subsection~\ref{uncertinty-analysis}, we design a MobileNetV2-based model to capture the aleatoric uncertainty~\cite{kendall2017uncertainties} of different regularization methods. Next, we set up an experiment to evaluate the models' robustness to random perturbations and out-of-distribution samples in subsection~\ref{robustness-analysis}. Finally, we study the transferability of the SIGN-generated samples to other models in subsection~\ref{transferibilty-analysis}.

\subsection{Classification Performance Analysis}\label{classification-performance}

In this subsection, we study the performance of the proposed approach and compare it with other methods in terms of the classification accuracy. We first test all the three regularization methods on the CIFAR-10 dataset using the MobileNetV2 and the BasicCNN models. As reported in Table~\ref{tab:classification}, the proposed method achieves the highest overall classification accuracy for both the models. Note that for the BasicCNN model, we apply the learned augmentation policies from normal CIFAR-10 images to both normal and SIGN transformed samples. We do this to study the transferability of the augmentation policies on the SIGN samples.

\begin{figure}[h!]
\centering
\includegraphics[width=\linewidth]{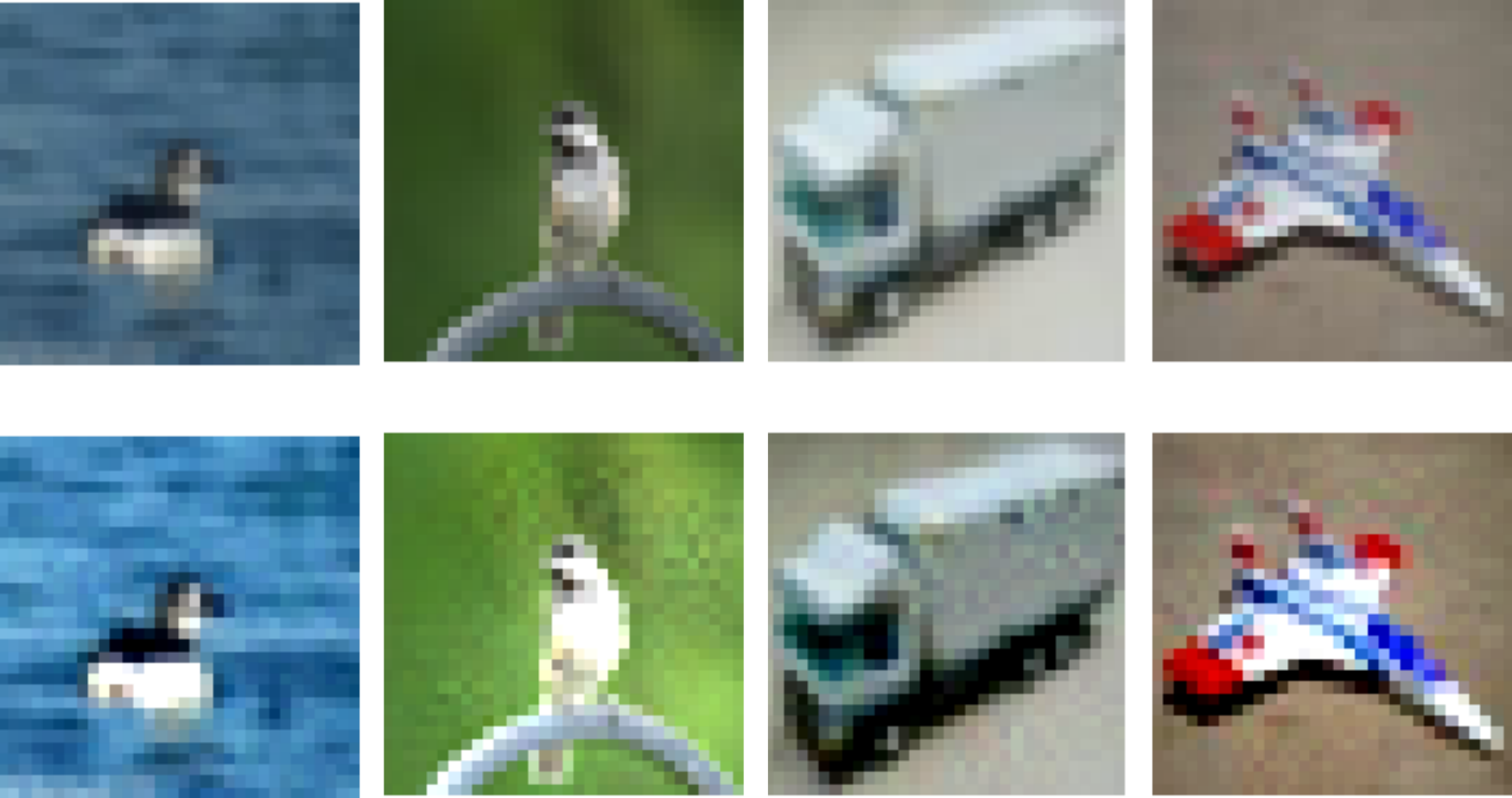}
\caption {CIFAR-10 samples before (top) and after (bottom) applying SIGN.}
\label{fig:cifar-sign}
\end{figure}

In Figure~\ref{fig:cifar-sign}, we visualize a few samples from CIFAR-10 before and after applying the SIGN regularization method. As can be seen, the color contrasts and the boundaries of the objects are enhanced after applying SIGN.

\begin{figure}[h!]
\centering
\includegraphics[width=\linewidth]{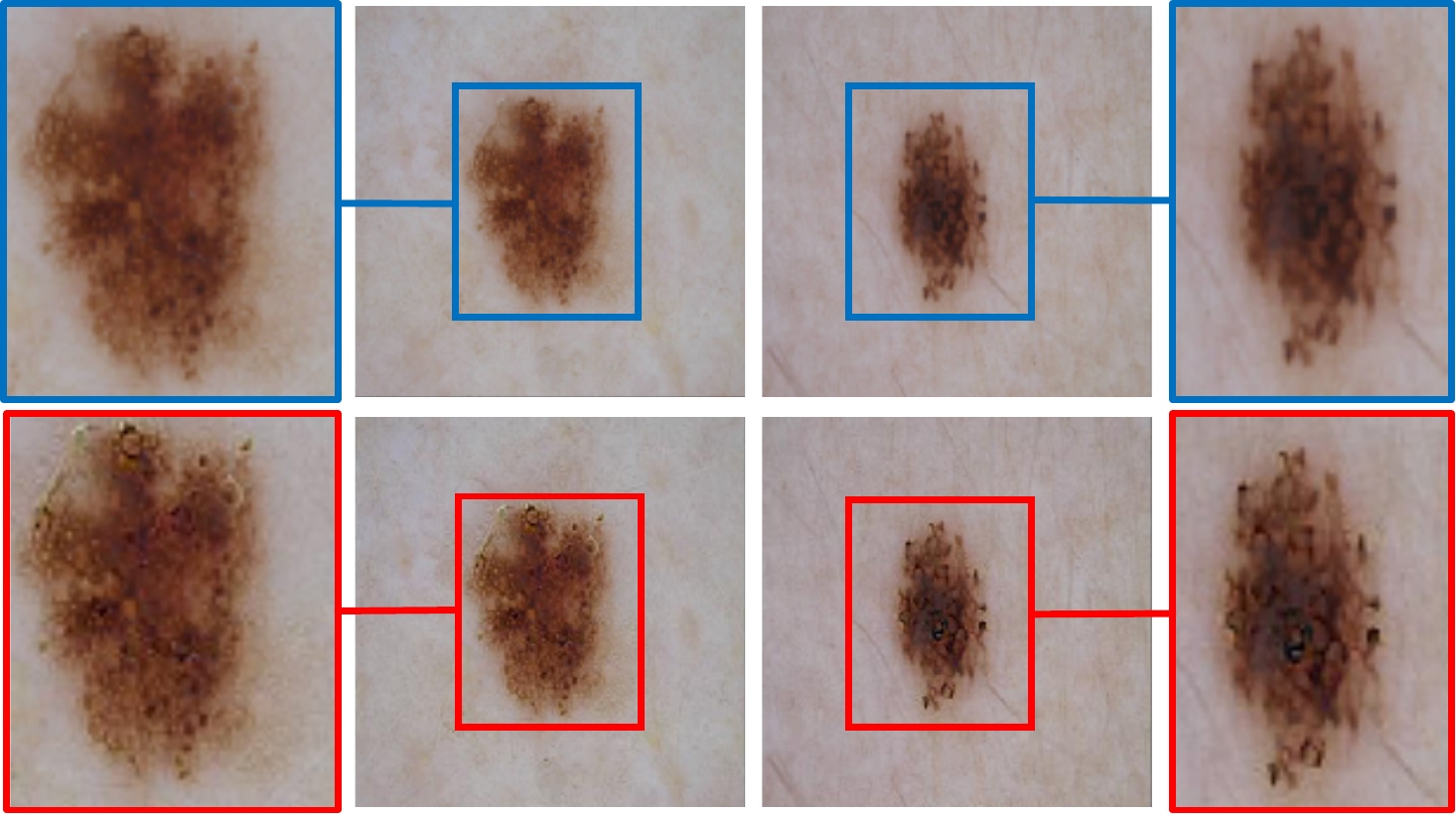}
\caption {ISIC skin lesion samples before (top) and after (bottom) applying SIGN.}
\label{fig:skin}
\end{figure}

Next, we apply the winner strategy (SIGN) from the previous experiments on a separate dataset and a model, i.e., the ISIC dataset and the Inception-ResNet-v2 model. The task here is to predict whether a skin lesion is a melanoma or not. The dataset consists of 2000 dermoscopic images of skin lesions, out of which only 374 belong to the positive class, indicating a considerable class imbalance. Our proposed SIGN method improves the baseline model's area under the curve (AUC) by $\sim 5\%$ and $ \sim 2\%$ on validation and test sets ($0.7633$ versus $0.8131$ and $0.7611$ versus $0.7787$), respectively. Note that for this experiment, we do not apply any data augmentation, so as to ascertain that the improvement is from the application of the SIGN method. It can be seen in Figure~\ref{fig:skin} that our SIGN method improves the dermoscopic features of the skin lesions, which leads to an improvement in the classification performance. 

\begin{figure*}[]
    \centering
        \begin{minipage}{.3\textwidth}
            \begin{subfigure}{\textwidth}
            \centering
            \includegraphics[width=\textwidth]{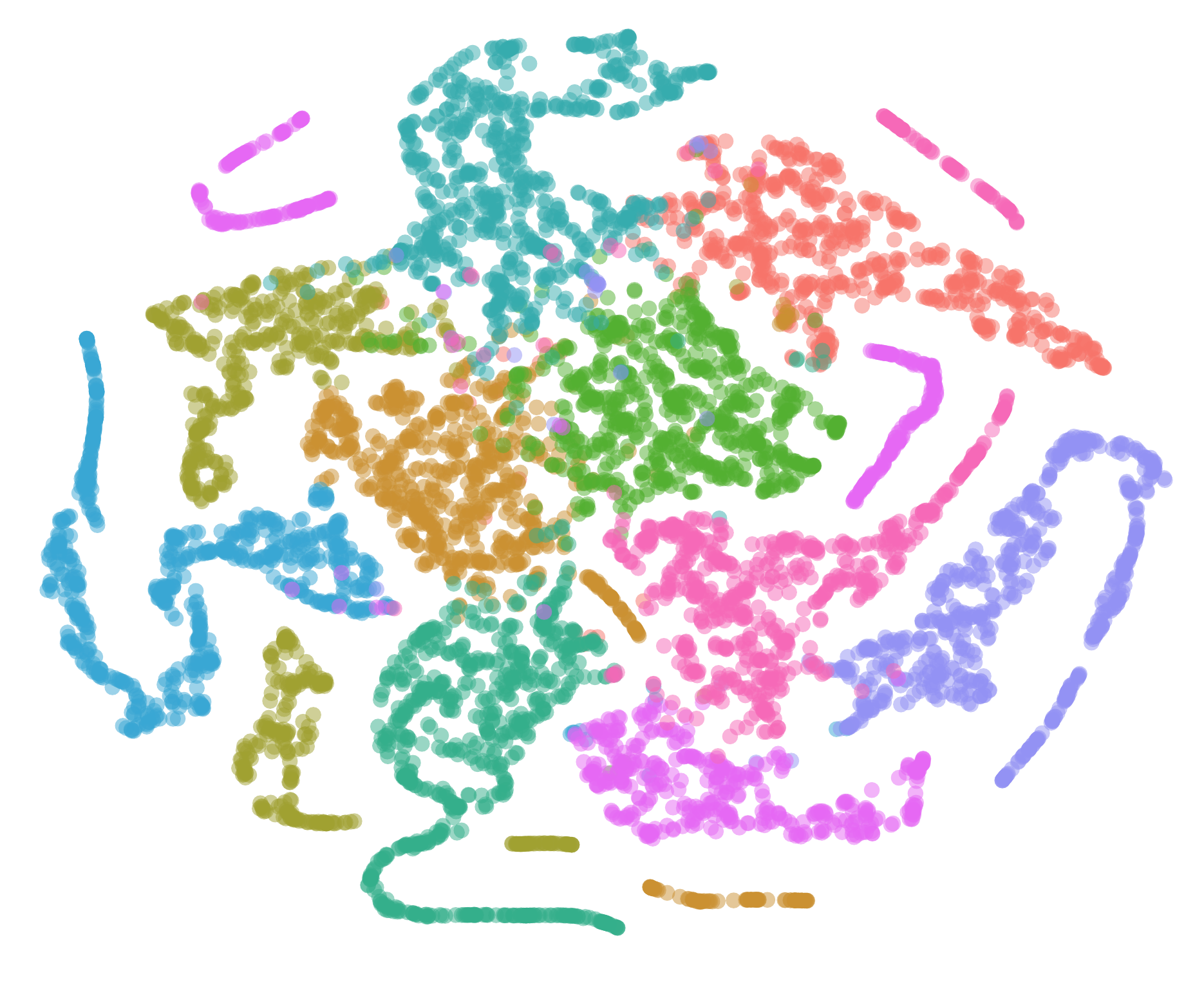}
            \end{subfigure}\\
            \begin{subfigure}{\textwidth}
            \centering
            \includegraphics[width=\textwidth]{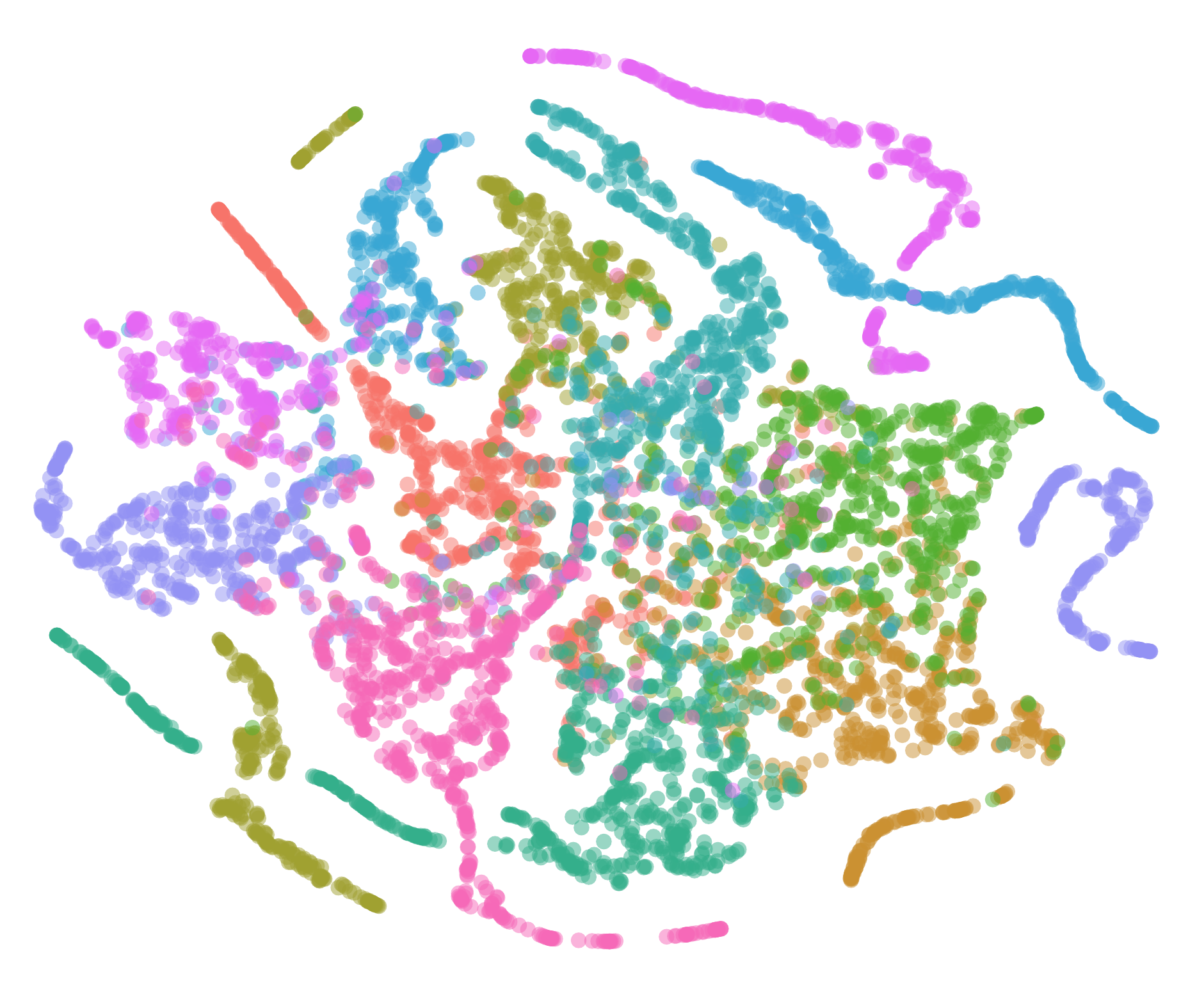}
            \caption{Classical~\cite{simonyan2014very}}
            \end{subfigure}%
        \end{minipage}
        \hfill
        \begin{minipage}{.3\textwidth}
            \begin{subfigure}{\textwidth}
            \centering
            \includegraphics[width=\textwidth]{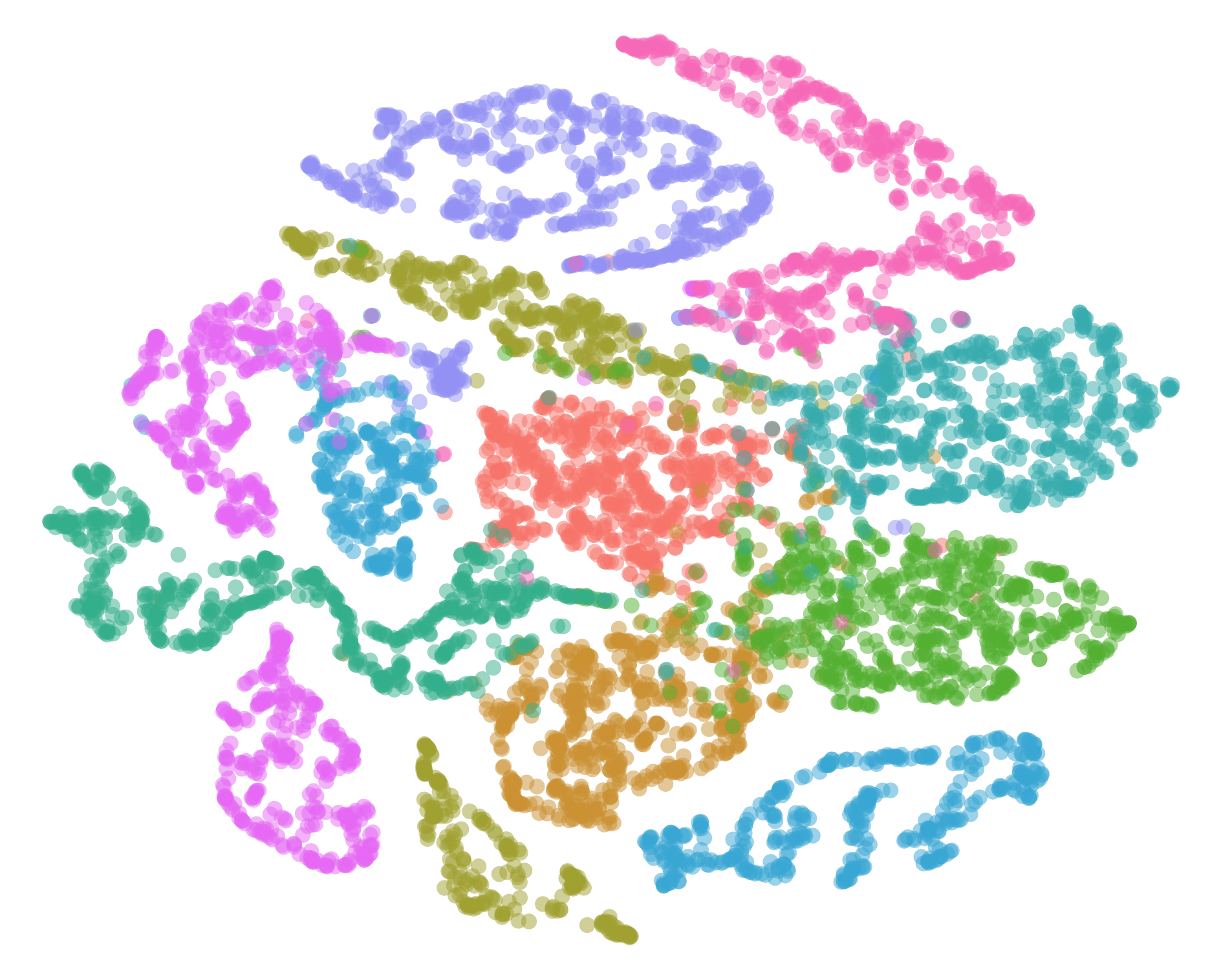}
            \end{subfigure}\\
            \begin{subfigure}{\textwidth}
            \centering
            \includegraphics[width=\textwidth]{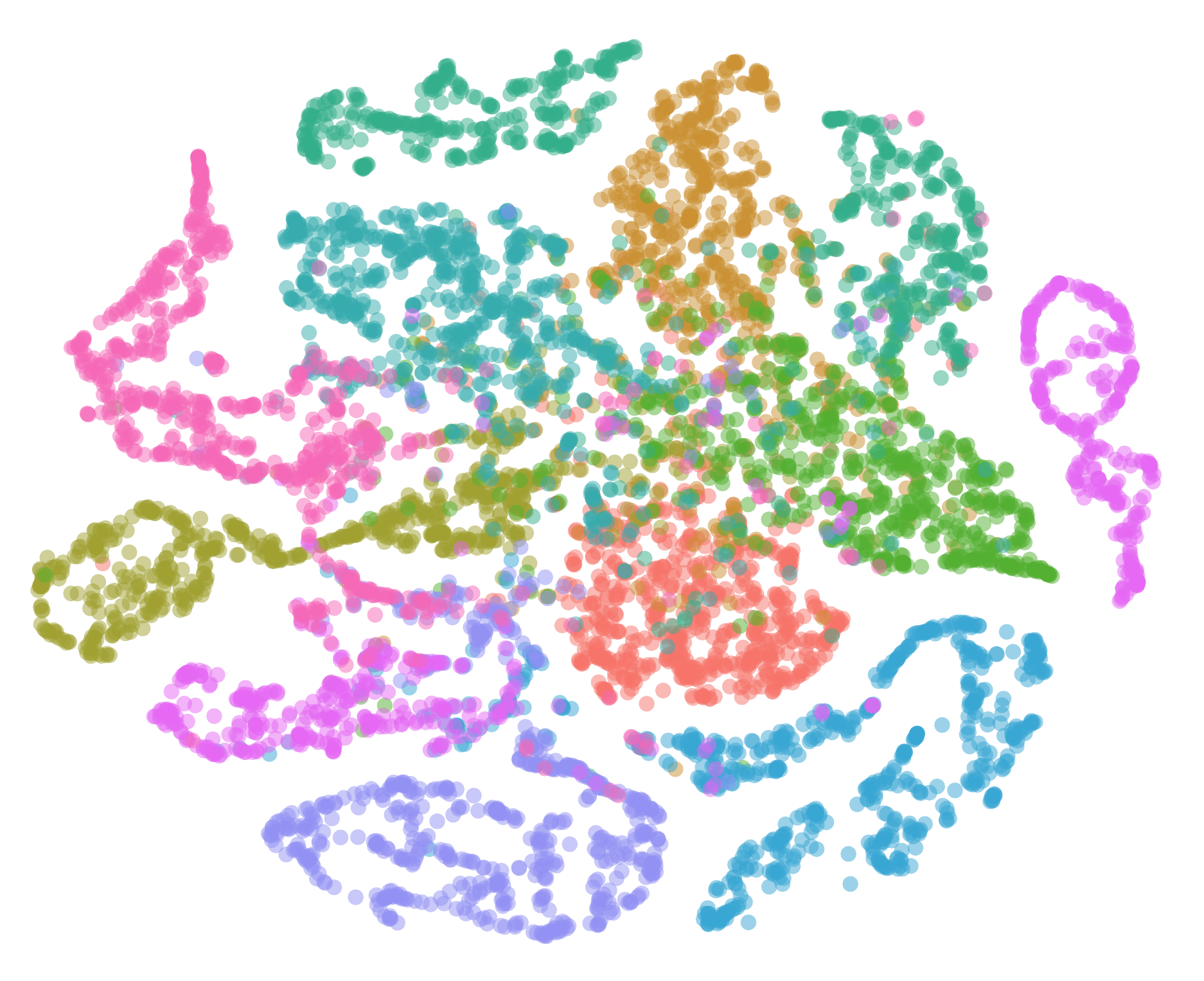}
            \caption{mixup~\cite{zhang2017mixup}}
            \end{subfigure}%
        \end{minipage}
        \hfill
        \begin{minipage}{.3\textwidth}
            \begin{subfigure}{\textwidth}
            \centering
            \includegraphics[width=\textwidth]{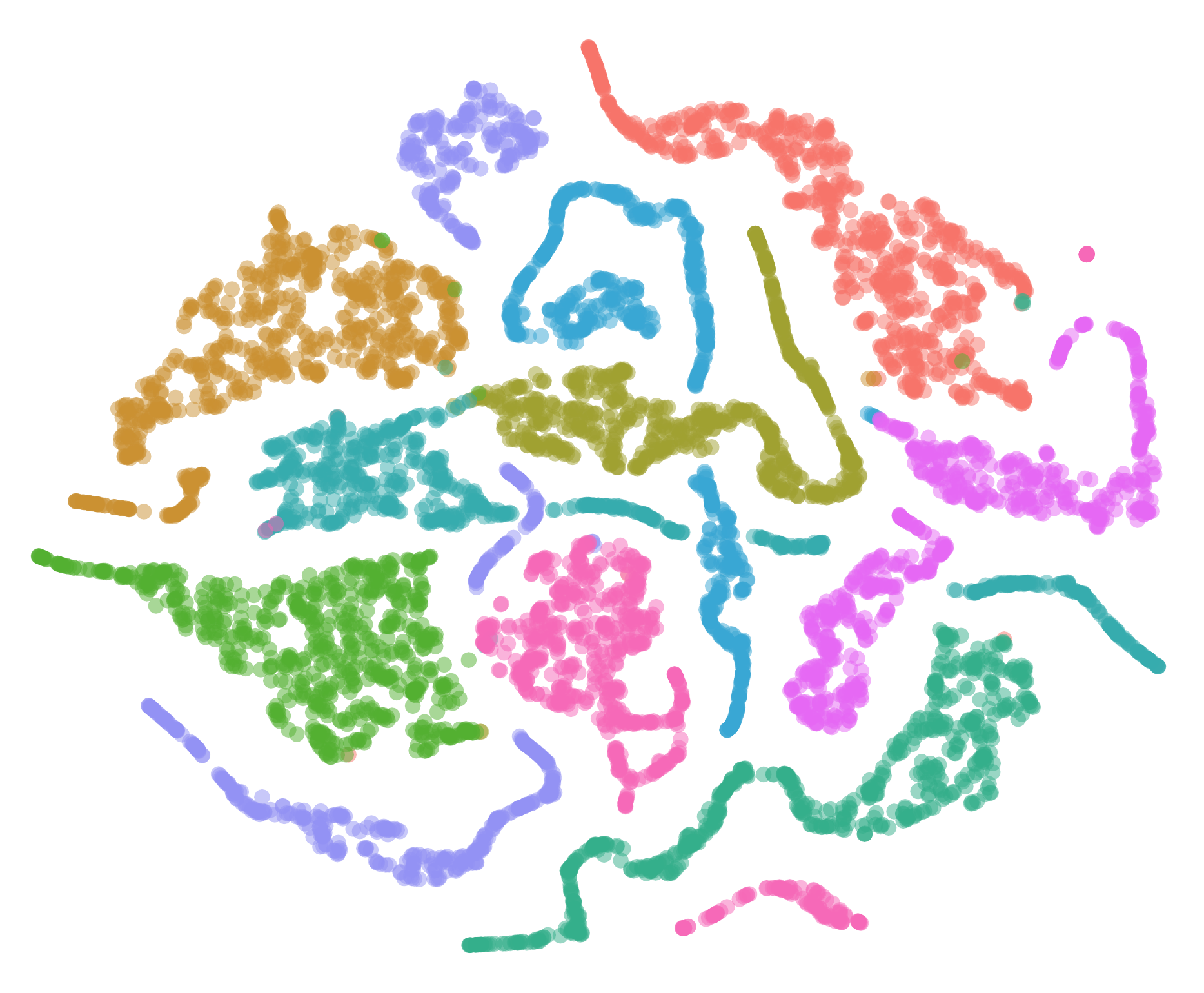}
            \end{subfigure}\\
            \begin{subfigure}{\textwidth}
            \centering
            \includegraphics[width=\textwidth]{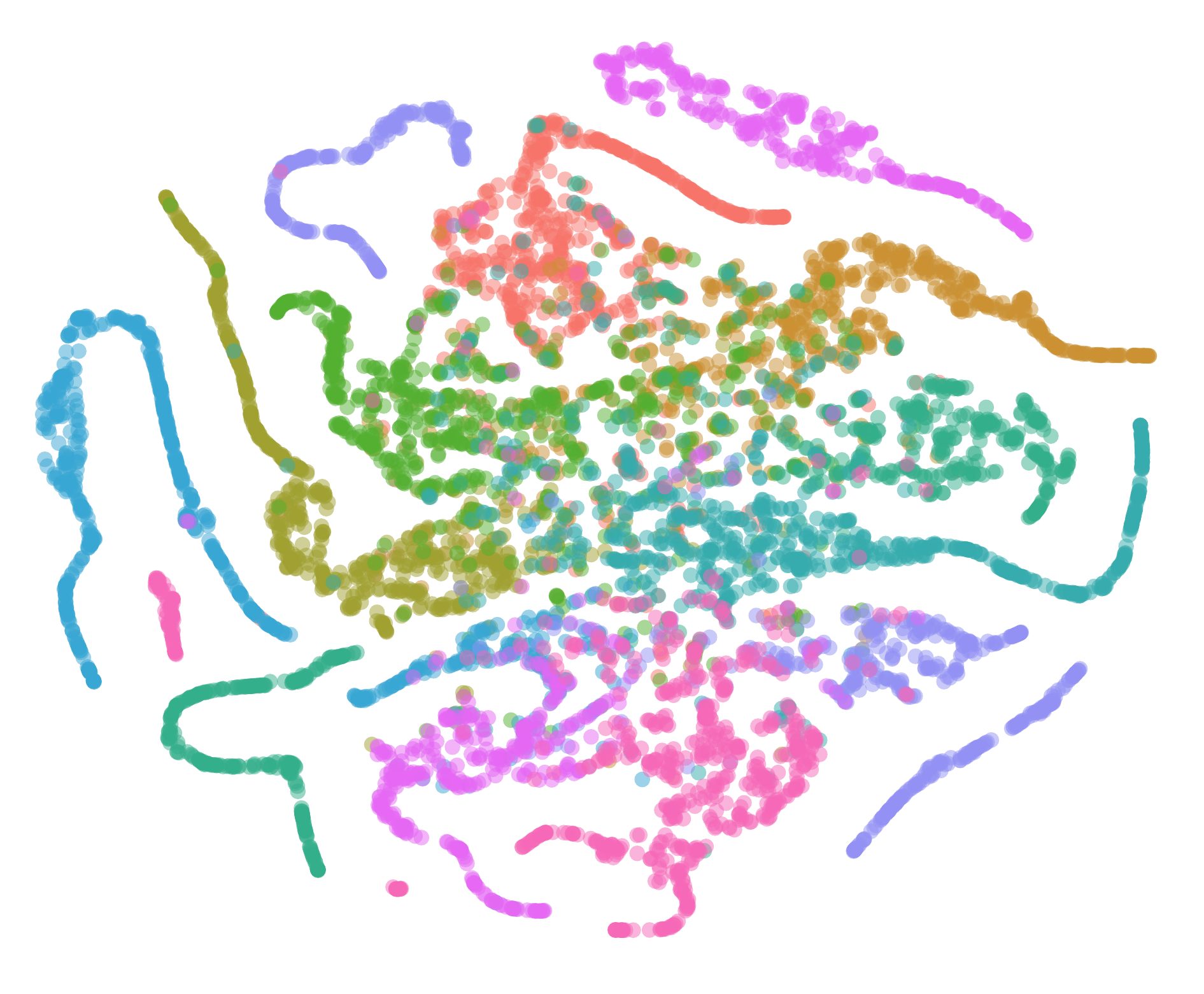}
            \caption{SIGN (proposed)}
            \end{subfigure}%
        \end{minipage}%
        \caption{Feature space visualization for models trained under different schemes. The first and second rows represent plots for training and validation sets, respectively.}
        \label{features}
\end{figure*}

Next, we visualize the t-SNE~\cite{maaten2008visualizing} feature space of the different approaches by transforming and projecting their high dimensional features into a 2D space. As can be seen in Figure~\ref{features}, the proposed SIGN method results in more compact class representations for both the training and the validation samples. However, for the mixup method, the linear interpolation function causes label smoothing that is manifested as fuzzy boundaries across the different classes.

In order to certify that there is useful signal for correct classification in the perturbation $\delta$ added by our iterative SIGN method to the samples, we train and evaluate a MobileNetV2 model with only the $\delta$s, i.e., (the linear estimation of the input) from the last layer of the model. We obtain a mean classification accuracy of 46.2\% for CIFAR-10 using only $\delta$s which shows that $\delta$ captures the classes information as expected. 

\subsection{Uncertainty Analysis} \label{uncertinty-analysis}
To capture uncertainties in the models' predictions, which arise because of the inherent noise in the observations, we model the aleatoric uncertainty~\cite{kendall2017uncertainties}. Particularly, we define a stochastic loss function which models the aleatoric uncertainty as

\begin{equation}
    \mathcal{L}_{x}=\sum_{i} \log \frac{1}{T} \sum_{t} \exp \left(\hat{x}_{i, t, c}-\log \sum_{c^{\prime}} \exp \hat{x}_{i, t, c^{\prime}}\right),
\end{equation}


\noindent where $c$ is the class label for sample input $i$. We design the model architecture such that it outputs $\mathbf{f}_{i}^{\mathbf{W}}$ and $\sigma_{i}^{\mathbf{W}}$ with parameters $\mathbf{W}$, where $\mathbf{f}_{i}$ is unary for input $i$ and $\hat{\mathbf{x}}_{i, 1\cdots t}$ are the $t$ Monte Carlo approximations (samples) of the unaries using the learned $\sigma_{i}^{\mathbf{W}}$ which is defined as 

\begin{equation}
    \hat{\mathbf{x}}_{i, t}=\mathbf{f}_{i}^{\mathbf{W}}+\sigma_{i}^{\mathbf{W}} \epsilon_{t}, \quad \epsilon_{t} \sim \mathcal{N}(0, I)
\end{equation}

\smallskip

Table~\ref{tab:classificationalea} summarizes the results of this experiment on CIFAR-10 dataset. As expected, the mixup method makes the decision boundaries smooth (see Figure~\ref{features}), which causes the model(s) trained with this data augmentation strategy to be highly under-confident, i.e., failing to know when it fails. In particular, the mixup method's threshold for correctly predicted classes is slightly above $0.1$, while the same value is $0.26$ and $0.33$ for the standard data augmentation and SIGN methods, respectively. This means that for a model trained with mixup, the lowest probability an image could have while still being correctly predicted is a little over $0.1$, which is just above the probability of randomly assigning a correct label for CIFAR-10 images.

\begin{table*}[!htbp]
\centering
\caption{Out-of-distribution results on the Tiny ImageNet dataset using a MobileNetV2 model trained on the CIFAR-10 dataset.}
\begin{tabular}{lccccccc}
\hline

Method  & Metric  & automobile  & bird  & cat  & dog  & frog  & Mean  \\ 
& (total images) & (49) & (146) & (50) & (298) & (100) & (643)\\ \hline
\multirow{3}{*}{Classical~\cite{simonyan2014very}}  & Accuracy  & $0.8163$  & $\textbf{0.5479}$  & $0.6800$  & $0.3289$  & $\textbf{0.8300}$  & $0.5210$ \\ \cline{2-8}

& \begin{tabular}[c]{@{}c@{}}$p \leq 0.5$\\(\# images)  \end{tabular}  & \begin{tabular}[c]{@{}c@{}}$0.4423$\\$\textbf{(1)}$  \end{tabular}  & \begin{tabular}[c]{@{}c@{}}$0.4225$\\$(12)$  \end{tabular}  & \begin{tabular}[c]{@{}c@{}}$\textbf{0.4438}$\\$(2)$  \end{tabular}  & \begin{tabular}[c]{@{}c@{}}$0.4135$\\$(17)$  \end{tabular}  & \begin{tabular}[c]{@{}c@{}}$0.4701$\\$(3)$  \end{tabular}  & \begin{tabular}[c]{@{}c@{}}$0.4240$\\$(35)$  \end{tabular}  \\

& [uncertainty]  & $[5.9622]$  & $[6.0362]$  & $[8.0968]$  & $[7.0100]$  & $[7.4542]$  & $[6.7464]$  \\ \hline

\multirow{3}{*}{mixup~\cite{zhang2017mixup}}  & Accuracy  & $0.7551$  & $0.4726$  & $0.6800$  & $0.2919$  & $0.7900$  & $0.4759$  \\ \cline{2-8}

& \begin{tabular}[c]{@{}c@{}}$p \leq 0.5$\\(\# images)  \end{tabular}  & \begin{tabular}[c]{@{}c@{}}$\textbf{0.4759}$\\$(3)$  \end{tabular}  & \begin{tabular}[c]{@{}c@{}}$\textbf{0.4314}$\\$(9)$  \end{tabular}  & \begin{tabular}[c]{@{}c@{}}$0.4133$\\$(6)$  \end{tabular}  & \begin{tabular}[c]{@{}c@{}}$0.4250$\\$(20)$  \end{tabular}  & \begin{tabular}[c]{@{}c@{}}$0.3867$\\$(3)$  \end{tabular}  & \begin{tabular}[c]{@{}c@{}}$\textbf{0.4256}$\\$(41)$  \end{tabular}  \\

& [uncertainty]  & $[4.8710]$  & $[4.5847]$  & $[4.4742]$  & $[4.8388]$  & $[3.8331]$  & $[4.6584]$  \\ \hline

\multirow{3}{*}{SIGN (proposed)}  & Accuracy  & $\textbf{0.9184}$  & $0.3973$  & $\textbf{0.7200}$  & $\textbf{0.4497}$  & $0.8000$  & $\textbf{0.5490}$  \\ \cline{2-8}

& \begin{tabular}[c]{@{}c@{}}$p \leq 0.5$\\(\# images)  \end{tabular}  & \begin{tabular}[c]{@{}c@{}}$0.3854$\\$\textbf{(1)}$  \end{tabular}  & \begin{tabular}[c]{@{}c@{}}$0.3125$\\$\textbf{(1)}$  \end{tabular}  & \begin{tabular}[c]{@{}c@{}}\textbf{N/A}\\$\textbf{(0)}$  \end{tabular}  & \begin{tabular}[c]{@{}c@{}}$\textbf{0.4611}$\\$\textbf{(3)}$  \end{tabular}  & \begin{tabular}[c]{@{}c@{}}\textbf{N/A}\\$\textbf{(0)}$  \end{tabular}  & \begin{tabular}[c]{@{}c@{}}$0.4162$\\$\textbf{(5)}$  \end{tabular}  \\

& [uncertainty]  & $[4.8829]$  & $[4.1209]$  & [N/A]  & $[6.3664]$  & [N/A]  & $[5.6206]$  \\ \hline

\end{tabular}
\label{tab:ood}
\end{table*}

\begin{table*}[h!]
\centering
\caption{Transferability results of the proposed SIGN method. Evaluating a NASNetMobile model trained using CIFAR-10 samples transformed using SIGN method from a MobileNetV2 model.}
\resizebox{\textwidth}{!}{%
\begin{tabular}{lccccccccccc}
\hline

Method  & airplane  & automobile  & bird  & cat  & deer  & dog  & frog  & horse  & ship  & truck  & Mean  \\ \hline
{Classical~\cite{simonyan2014very}}  & $0.846$  & $0.873$  & $0.717$  & $0.620$  & $0.722$  & $\textbf{0.741}$  & $\textbf{0.962}$  & $0.804$  & $0.907$  & $0.893$  & $0.8085$  \\ \hline

{Transfer (with SIGN)}  & $\textbf{0.858}$  & $\textbf{0.927}$  & $\textbf{0.813}$  & $\textbf{0.672}$  & $\textbf{0.829}$  & $0.740$  & $0.897$  & $\textbf{0.898}$  & $\textbf{0.934}$  & $\textbf{0.910}$  & $\textbf{0.8478}$  \\ \hline

\end{tabular}
}
\label{nasnet-transfer}
\end{table*}

\begin{table*}[!htbp]
\centering
\small
\caption{Robustness of the methods to additive noise and pixel corruption.}
\begin{tabular}{llccc}
\hline
    \multirow{2}{*}{Model}  & \multirow{2}{*}{Method}  & \multirow{2}{*}{Normal accuracy}   & Pixels turned off                       & Gaussian noise            \\ 
       & & & (50 random pixels) & ($\mu=0,\sigma=10$) \\ \hline
    \multirow{3}{*}{MobileNetV2~\cite{sandler2018mobilenetv2}} 
    & Classical~\cite{simonyan2014very} & $0.8550$ & $0.6672 \pm \num{3e-4}$ & $0.6552 \pm \num{4e-5}$ \\
    & mixup~\cite{zhang2017mixup} & $0.8584$ & $0.5970 \pm \num{2e-4}$ & $0.6590 \pm \num{9e-5}$ \\
    & SIGN (proposed) & $\textbf{0.8637}$ & $\textbf{0.6690} \pm \textbf{3e--4}$ & $\textbf{0.7122} \pm \textbf{3e--5}$ \\ 
    \hline
    \multirow{2}{*}{BasicCNN (Section~\ref{sec:experiments})} 
    & DeepAugment~\cite{deepaugment} & $0.8648$ & $0.8361 \pm \num{1e-5}$ & $0.7875 \pm \num{1e-5}$ \\
    & DeepAugment + SIGN & $\textbf{0.8691}$ & $\textbf{0.8479} \pm \textbf{4e--6}$ & $\textbf{0.7980} \pm \textbf{2e--5}$ \\ 
    \hline
\end{tabular}
\label{corr}
\end{table*}

In Table~\ref{tab:classificationalea}, we report the mean probability values of the correctly classified test samples with a correct class probabilities less than $0.5$, i.e., relatively low confidence. As can be seen in the table, for all the classes combined, our proposed method only has $26$ samples in total with probabilities below the threshold value, while there are $363$ and $212$ samples for mixup and the standard data augmentation methods respectively. As can be seen in the last row of the same table, the SIGN method obtains reasonably high uncertainty values for the samples with a low confidence level, which is desirable, since we want a model to know when it fails by producing high uncertainty. However, the mixup method causes the model to produce relatively smaller uncertainty values while it is still highly under-confident, 
indicating that the model is unaware of its failures.

\subsection{Out-of-distribution and Robustness Analysis} \label{robustness-analysis}
\begin{figure}[h!]
\centering
\small
\includegraphics[width=\linewidth]{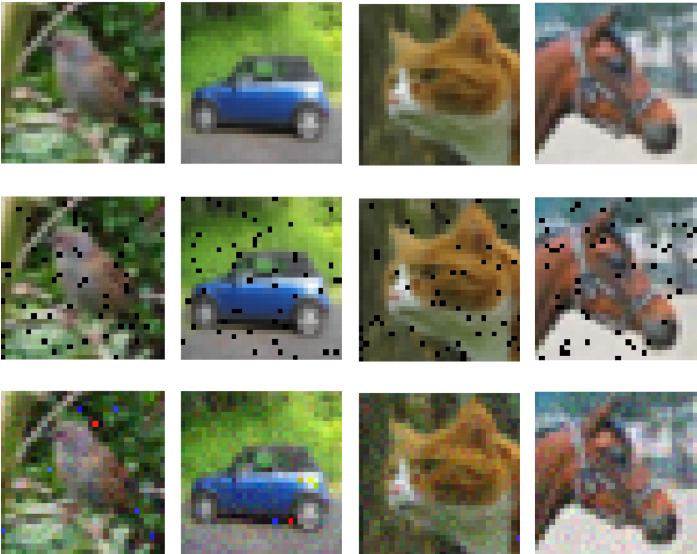}
\caption {Samples of the corrupted CIFAR-10 images. Rows from top to bottom show the normal, pixel-off, and Gaussian noise, respectively.}
\label{rbst}
\end{figure}

Next, we study the performance of the models trained with different data augmentation techniques by evaluating on out-of-distribution and corrupted samples, a few examples of which are shown in Figure~\ref{rbst}. For evaluating on out-of-distribution samples, we train a MobileNetV2 model on the CIFAR-10 dataset and test on Tiny ImageNet. In particular, we extract images from multiple classes in the TinyImageNet dataset so as to correspond to the classes present in CIFAR-10 and resize them to the CIFAR-10 image resolution ($32\times32$). We use 49 images of `automobile' class, 146 images of `bird' class, 50 images of `cat' class, 298 images of `dog' class, and 100 images of `frog' class, totaling to 643 images overall. As can be seen in the third row of Table~\ref{tab:ood}, our method achieves significantly better results ($\sim 7\%$ improvement over mixup) compared to other techniques. 

Next, to test the robustness of the methods to different input corruptions using two different models and five different augmentation approaches, we apply two types of corruptions to the test samples ( which are assumed to be drawn from the same distribution as the train samples): (a) we add Gaussian noise with mean $0$ and standard deviation of $10$ to the images, and (b) we turn off 50 randomly selected pixels in each image. Table~\ref{corr} summarized the results for this experiment. The SIGN method shows higher resistance from other methods. The DeepAugment method's results are improved from $83.61\%$ to $84.79\%$ when it uses the SIGN transformed samples.  

This experiment shows that if a method is restricted to learn from a fewer input variables, as SIGN does, it should be robust when a few  variables are missing, which is simulated in this experiment by zeroing out random pixels.

\subsection{Transferability of SIGN Samples} \label{transferibilty-analysis}
Finally, we examine whether the transformed samples obtained with our SIGN method are effective in other unknown models. To this end, we leverage the samples mapped to a new space using a MobileNetV2 model to train a NASNetMobile model. As reported in Table~\ref{nasnet-transfer}, the SIGN method's samples are transferable as they improve the  NASNetMobile baseline model's mean classification accuracy on the CIFAR-10 dataset by $\sim 4\%$, i.e., from $0.8085$ to $0.8478$, including considerable improvements in class-wise accuracies for 8 out of 10 classes.


\section{Conclusion}
We proposed SIGN, a regularization method that can be used as a data augmentation strategy. Our proposed iterative SIGN technique produces different transformations of the input data, which are then used to train a model. We showed how the push-forward concept in manifold transformation could be applied to both obtain the linear estimation of the input using a layer in the model and diminish the effect of the non-important input variables by assigning them negative signs. We showed that the iterative SIGN method could help for better generalization performance of deep models. 

We also discussed the critical limitations and risks of using the mixup and the adversarial training methods as regularization techniques. We evaluated the proposed idea for several classification tasks and demonstrated the superior classification accuracy obtained using iterative SIGN regularization. Moreover, we note that the proposed method can also be extended to other problems such as dense image labeling and object detection. Another possible future direction is to study the feasibility of using the SIGN method to cancel out the adversarial perturbations added by adversarial attack strategies.

{\small
\bibliographystyle{ieee}
\bibliography{egpaper_final}
}


\end{document}